\documentclass{article}

\RequirePackage{snapshot}

\usepackage[preprint,nonatbib]{neurips_2019}

\usepackage[utf8]{inputenc} 
\usepackage[T1]{fontenc}    
\usepackage{hyperref}       
\usepackage{url}            
\usepackage{booktabs}       
\usepackage{amsfonts}       
\usepackage{nicefrac}       
\usepackage{microtype}      

\usepackage{comment}
\usepackage{amsmath}
\usepackage{color}
\usepackage{algorithm2e}
\usepackage{graphicx}

\usepackage{cuted}
\usepackage{subcaption}

\newcommand{\figref}[1]{Fig.~\ref{fig:#1}}

\title{Object as Distribution}

\author{%
  Li Ding \\
  MIT \\
  \texttt{liding@mit.edu} \\
  \And
  Lex Fridman \\
  MIT \\
  \texttt{fridman@mit.edu} \\
}

\begin{document}

\maketitle

\begin{abstract}
  Object detection is a critical part of visual scene understanding. The representation of the object in the detection
  task has important implications on the efficiency and feasibility of annotation, robustness to occlusion, pose,
  lighting, and other visual sources of semantic uncertainty, and effectiveness in real-world applications (e.g.,
  autonomous driving). Popular object representations include 2D and 3D bounding boxes, polygons, splines, pixels, and
  voxels. Each have their strengths and weakness. In this work, we propose a new representation of objects based on the
  bivariate normal distribution. This distribution-based representation has the benefit of robust detection of
  highly-overlapping objects and the potential for improved downstream tracking and instance segmentation tasks due to
  the statistical representation of object edges. We provide qualitative evaluation of this representation for the
  object detection task and quantitative evaluation of its use in a baseline algorithm for the instance segmentation
  task.
\end{abstract}


\section{Introduction}

Object detection has served as one of the defining problems in the field of computer vision for over 50 years
\cite{papert1966summer}. The ``object'' with its relation to the scene has no universal formalization or definition -- a
topic under extensive research and debate in mathematics, computer science, cognitive science, and philosophy. With
every attempt at explicitly defining what it means to be a distinct object in the visual scene, a lot of valuable
semantic knowledge is discarded \cite{feldman2003visual}. In computer vision, objects in 2D image space have been
defined by their 2D bounding boxes \cite{redmon2016you}, 3D bounding boxes \cite{mousavian20173d}, polygons
\cite{castrejon2017annotating}, splines \cite{castro2015statistical}, pixels \cite{liang2018proposal}, and voxels
\cite{girdhar2016learning}. Each representation has benchmarks and state-of-the-art algorithms. Each have strengths and
weaknesses when considered from the pragmatic perspective of a particular application (e.g., robot vision), providing
variable levels of fidelity, information density, and annotation cost.

We propose a new representation based on the bivariate normal distribution (5 parameters) as an alternative to the most
commonly used object representation of 2D bounding boxes (4 parameters). This distribution-based representation has the
benefit of robust detection of highly-overlapping objects as shown in \figref{title}. No well-established benchmarks are available to evaluate this
statistical representation, so for the detection task we rely primarily on qualitative evaluation. Conceptually, the
strength of this representation is its emphasis on the object center of visual mass versus the object edges, allowing
for uncertainty around the latter. The result of this emphasis is that the derived tasks of object tracking and instance
segmentation may become more robust to the inherent spatial and temporal variability of object edges and to occlusion
artifacts. We provide a baseline instance segmentation approach based on this statistical representation, motivating
further work on utilizing this representation for the downstream segmentation and tracking tasks. Ultimately, object
detection is a simplification of the general task of perception and visual scene understanding. One of the underlying
open questions raised by this work is whether bounding boxes is the most useful minimalist representation of objects in
real-world detection tasks.

\newcommand{\myfig}[3]{%
  \begin{subfigure}[t]{#2\textwidth}
    \includegraphics[width=\textwidth]{images/title/#1.png}%
    \caption{#3}
    \label{fig:title_#1}
  \end{subfigure}
}
\vspace{0.2in}
\newcommand{\myspace}{\hspace{0in}}
\captionsetup[sub]{labelformat=parens}
\begin{figure}
  \centering
  \captionsetup{type=figure}
  \myfig{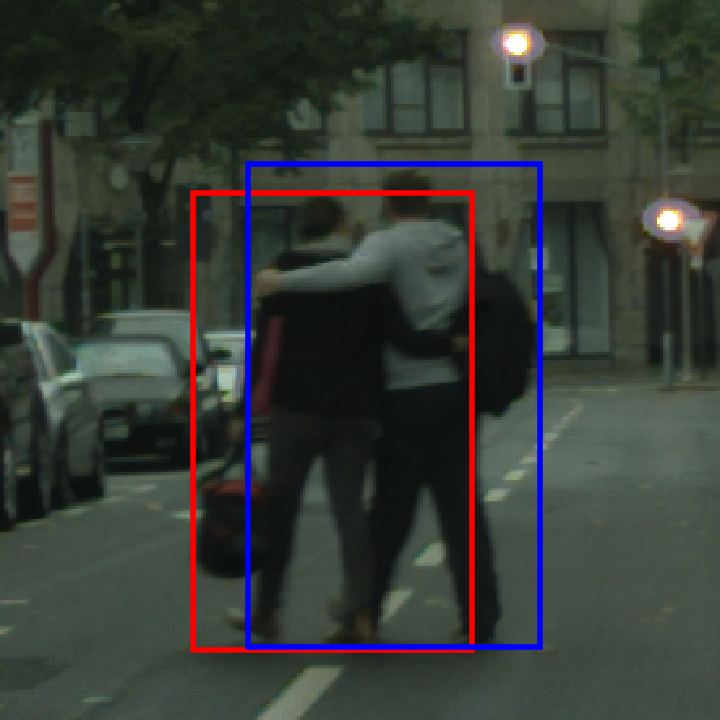}{0.32}{Bounding box representation.}\myspace
  \myfig{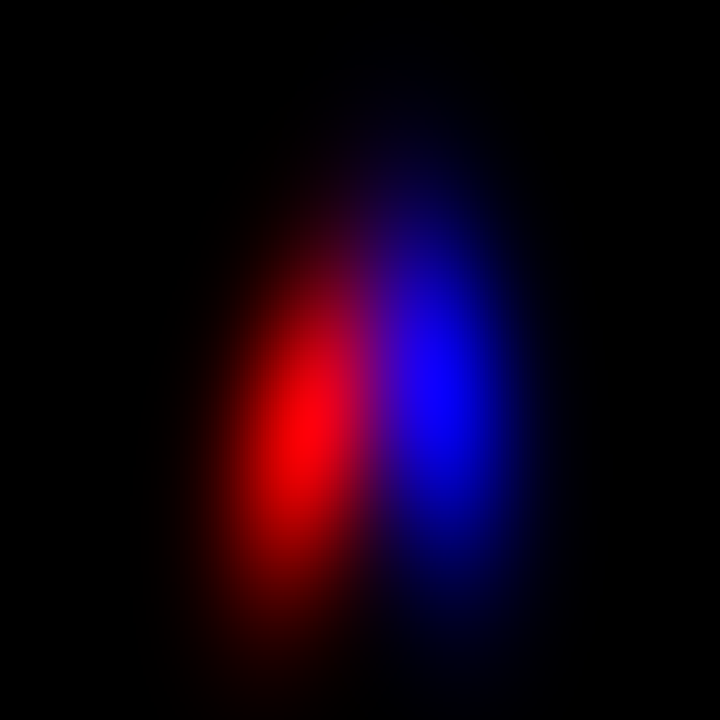}{0.32}{Distribution representation.}\myspace
  \myfig{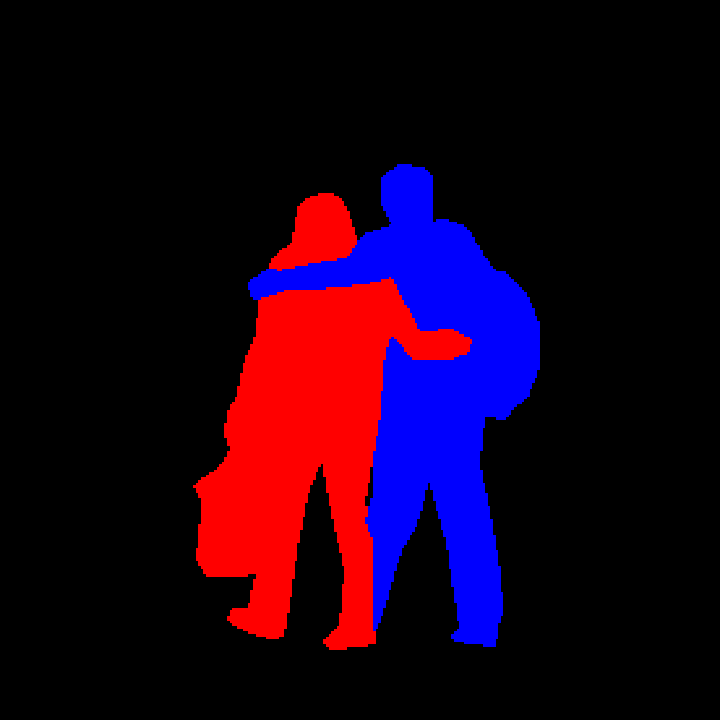}{0.32}{Pixel representation.}
  \caption{Illustrative example of object representations considered in this work, highlighting the case of
    highly-overlapping objects and the robustness of each representation to the task of successfully decoupling the
    objects in the detection step.}
  \label{fig:title}
\end{figure}
\let\myfig\undefined
\let\myspace\undefined

\section{Related Work}

\subsection{Object Detection}

Convolutional neural networks (CNNs) have been used in recent years to achieve state-of-the-art performance on object
detection \cite{girshick2014rich,ren2015faster,liu2016ssd}. These CNN-based methods can be divided into two types: one
stage methods and two stage methods. One stage methods such as YOLO \cite{redmon2016you} or SSD \cite{liu2016ssd} which
directly predict the bounding boxes of interest with a single feedforward pass through the nextwork. Two stage methods
such as Faster R-CNN \cite{ren2015faster} or R-FCN \cite{dai2016r} first generate proposals, and then exploit the
extracted region features from CNN for further refinement. Further refinement of these methods focus on addressing
various drawbacks such as the lack of robustness to scale variation, often achieving state-of-the-art performance on
object detection benchmarks \cite{li2019scale} (e.g., on COCO Object Detection Task \cite{lin2014microsoft}).

\subsection{Box-Free Instance Segmentation}

Although instance segmentation has been viewed as a more advanced form of object
detection, many recent advances for instance segmentation are still
box-dependent, e.g. \cite{dai2016instance, li2017fully, he2017mask,
arnab2017pixelwise} involves first detecting objects with a box and then
segmenting each object using the box as a guide, \cite{pinheiro2015learning,
dai2016instances} generate mask proposals in a dense sliding-window manner. On
the other hand, box-free methods~\cite{bai2017deep, kirillov2017instancecut,
liu2017sgn} predict each image pixel with a class label and some auxiliary
information, then use a clustering algorithm to group pixels into object
instances. A major drawback for these methods is that the auxiliary information
is usually uninterpretable, so the detection results can not be obtained until
the dense object masks are generated, which is sometimes unnecessary and
expensive to compute.

\section{Rethinking Object Representation}

One of the basic component in object detection is the way of representing the
existence of objects of interest in the space and time. In the case of 2D image,
the axis-aligned minimum bounding box representation is a wildly used method to
identify an object with its approximate location and a very simple box
descriptor of its shape and size. Despite its advantages such as
parameterization- and annotation-friendly, to be used as the label to denote the
existence of an object, bounding box has a few drawbacks such as: 1) it only
cares about the borders of the object on two directions, which is not
representative if the object is rotated or not squarely shaped. 2) it is
sensitive to the change on the border parts, meaning the box parameters may
change dramatically ignoring the majority pixels of the object. 3) It can not
distinguish overlapped objects within very similar bounding boxes, though their
masks may be very different.

On the other hand, another representation of 2D object that has been recently
used is the object mask, usually parameterized by a dense pixel matrix or a
polygon. It gives highly accurate shape of the object, but usually have too many
parameters that makes it hard to be used as the label. An example is in
state-of-the-art proposal-based instance segmentation methods, e.g. Mask
R-CNN~\cite{he2017mask}, that involve first detecting each object with a
bounding box then segmenting it, the box is still responsible for denoting the
existence of an object. 

In the consideration of the above two paradigms, our goal is to explore a new
way of representing the existence of 2D object, that can at most describe the
object, making it distinguishable from other objects, with the least number of
parameters.

\subsection{2D Object Representation with Bivariate Normal Distribution}
\label{subsec:representation}

We hereby introduce a representation, that uses bivariate normal distribution to
parameterize the visual existence of 2D objects in the scene. Specifically, for
any densely or coarsely annotated object $i$ in the image, its annotation can be
viewed as a set of pixels $\mathbb{Z}^{(i)}$ distributed in the 2D space. We use
the bivariate normal distribution to represent this set as

\begin{equation}
  \mathbb{Z}^{(i)} \sim \mathcal{N}_2(\boldsymbol{\mu}_i, \boldsymbol{\Sigma}_i)
\end{equation}

\begin{equation}
 \boldsymbol{\mu}_i = \mathrm{E}\bigl[
 \mathrm{E}[\mathbf{X}_i],\mathrm{E}[\mathbf{Y}_i]
 \bigr]^{\mathsf{T}}
\end{equation}

\begin{equation}
 \boldsymbol{\Sigma}_i = 
 \begin{bmatrix}
 \mathrm{E} [ (\mathbf{X}_i - \mathrm{E}[\mathbf{X}_i])^2] &
  \mathrm{E} [ (\mathbf{X}_i - \mathrm{E}[\mathbf{X}_i])(\mathbf{Y}_i - 
 \mathrm{E}[\mathbf{Y}_i]) ] \\
 \mathrm{E} [ (\mathbf{X}_i - \mathrm{E}[\mathbf{X}_i])(\mathbf{Y}_i - 
 \mathrm{E}[\mathbf{Y}_i]) ] &
  \mathrm{E} [(\mathbf{Y}_i - \mathrm{E}[\mathbf{Y}_i])^2 ]
 \end{bmatrix}
\end{equation}

where $\mathbf{X}_i$ and $\mathbf{Y}_i$ are the x-axis-coordinates and 
y-axis-coordinates of all the pixels belong to the object, respectively. 

By using maximum likelihood to estimation, we can parameterize the distribution as

\begin{equation}
\boldsymbol{\mu}_i =  [\mu_{x_i}, \mu_{y_i}]^{\mathsf{T}}
\end{equation}

\begin{equation}
\boldsymbol{\Sigma}_i = \begin{bmatrix}
\sigma_{x_i}^2 &
\rho_i\sigma_{x_i}\sigma_{y_i} \\
\rho_i\sigma_{x_i}\sigma_{y_i} &
\sigma_{y_i}^2
\end{bmatrix}
\end{equation}

where $\mu_{x_i}$, $\mu_{y_i}$, $\sigma_{x_i}$, $\sigma_{y_i}$ are means and
standard deviations of $\mathbf{X}_i$ and $\mathbf{Y}_i$, respectively. $\rho_i$
is the correlation coefficient of $\mathbf{X}_i$ and $\mathbf{Y}_i$. 

There are several advantages of using the above representation, comparing to the
2D bounding box representation. Firstly, the proposed representation is more
precise in terms of the shape and rotation variation of general objects,
especially for objects that are not squarely shaped. Secondly, it gives more
robust indication of object's existence in the 2D space, as opposed to boxes
that will vary significantly when there are any changes on the boarder of an
object. Thirdly, it can handle a few difficult situations such as distinguishing
highly-overlapped objects, which will be discussed in detail in the next
section. In addition, this representation is parameterized by only five
parameters for each object, one extra than the bounding box but much fewer than
the dense pixel labeling.

\subsection{Distinguishing Objects with Discrimination Information}

For any successful representation of a certain kind information, it is crucial
that the representation should minimize the information loss at the time of
recovery. For 2D objects in this case, the most important information to
preserve is the spatial existence of all annotated objects that also makes them
distinguishable from each other (one can not have same representation for two
different objects). It is also an important premise for many recent object
detection approaches, namely those who tend to output a smooth response map that
triggers many imprecise object hypotheses, that by using some post-processing
techniques, e.g. non-maximum suppression, one can ideally remove false positives
and obtain a best single detection for each object. 

Under the situation of using parameterized distributions to encode objects, a
natural thinking of distinguish them is to measure how one distribution is
different from another. In this work, we use the Kullback–Leibler (KL)
divergence $D_{\mathrm{KL}}(\mathbb{Z}^{(i)} \parallel \mathbb{Z}^{(j)})$
between two parameterized distributions $\mathbb{Z}^{(i)}$ and
$\mathbb{Z}^{(j)}$ to quantify the difference between two objects. Note that the
KL divergence is always non-negative, with $D_{\mathrm{KL}}(\mathbb{Z}^{(i)}
\parallel \mathbb{Z}^{(j)}) = 0$ if and only if $\mathbb{Z}^{(i)} =
\mathbb{Z}^{(j)}$ almost everywhere, which is statistically unlike to happen in
the natural scene. In practice, for model optimization and inference, we use the
symmetrified KL divergence $D$ as:
\begin{equation}
  \label{eq:kl}
  D_{\mathrm{KL}^2}(\mathbb{Z}^{(i)} \parallel \mathbb{Z}^{(j)}) = \frac12(D_{\mathrm{KL}}(\mathbb{Z}^{(i)} \parallel \mathbb{Z}^{(j)}) 
  + D_{\mathrm{KL}}(\mathbb{Z}^{(j)} \parallel \mathbb{Z}^{(i)}))
\end{equation}
in order to ensure the consistency of the measure of two objects with different ordering.

\begin{figure}
  \centering
  \includegraphics[width=.96\linewidth]{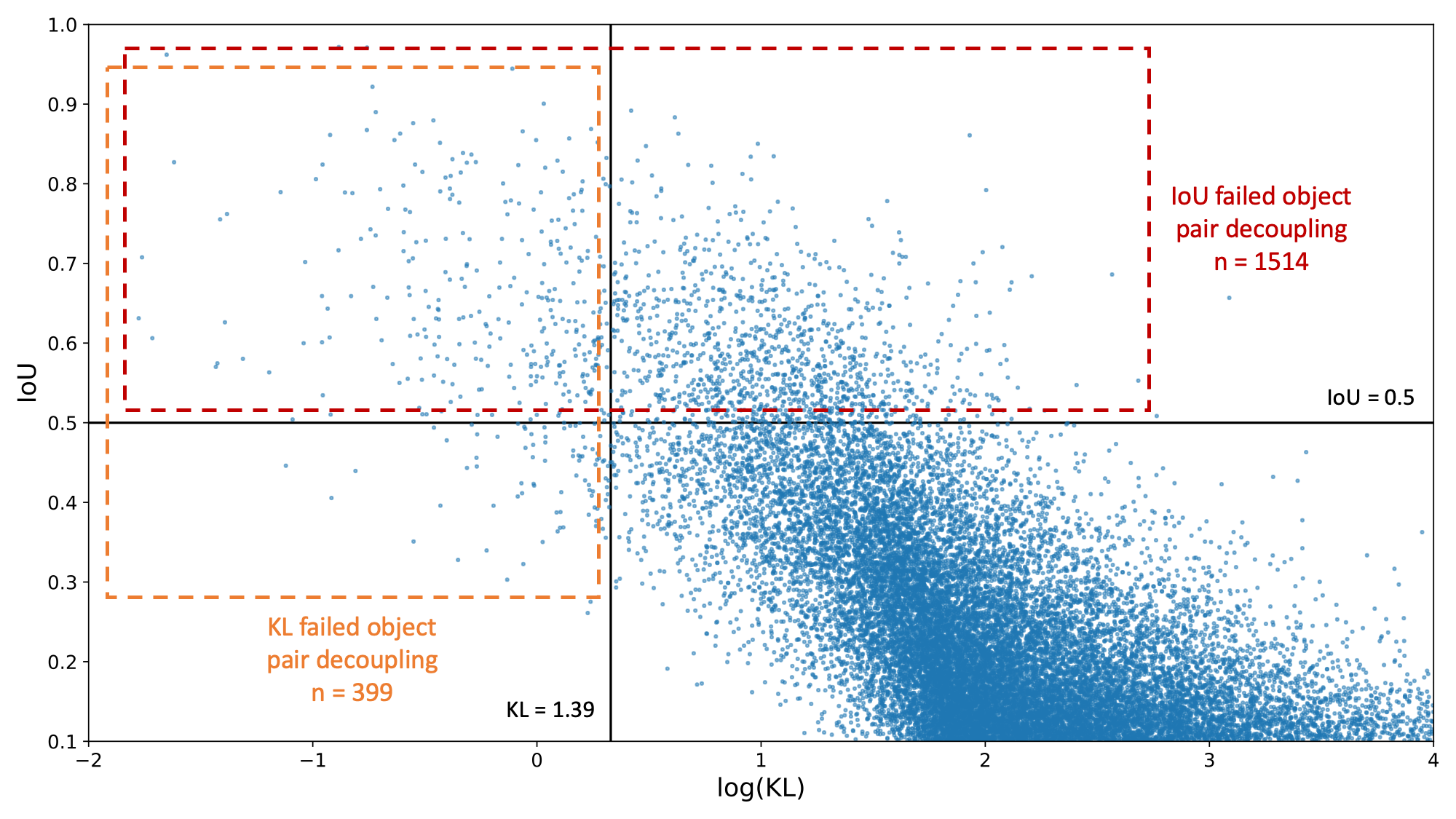}
  \caption{Comparison between KL Divergence and IoU on Cityscapes \textit{train}. Each dot is a pair of two objects in the same scene, measured by the KL divergence between their distribution representation (x-axis) and the bounding box IoU (y-axis). The horizontal line is the threshold of 0.5 IoU used by many detection models and evaluation metrics. The vertical line is the threshold we use to distinguish objects.}
  \label{fig:kliou}
\end{figure}

For object detection in the literature, intersection over union (IoU) is a major evaluation metric to measure the
difference between two detections with bounding box representation. Our proposed method of distribution representation
with KL divergence not only shares some good features with IoU including invariance to the scale of image and strictness
about both location and size of objects, but also has a few advantages. Firstly, KL Divergence is fully differentiable
and can be directly used in optimization. This eliminates one of the major drawbacks of IoU that people have to find
alternative ways to optimize bounding box size and location. Secondly, it is able to handle edge cases when the objects
are overlapped and have very similar bounding boxes. This situation is very likely to happen in some real-world
contexts, e.g. driving scene, where pedestrians or vehicles crowds are likely to contain highly overlapping
objects. Fig.~\ref{fig:kliou} shows some statistics obtained from the Cityscapes~\cite{cordts2016cityscapes} dataset,
which consists of various driving scenes in urban areas. Our representation significantly improve the discrimination of
highly-overlapped objects by reducing the number of failed object pair decoupling by over 70\%.

\section{Object Detection with Representation Modeling}

To further illustrate the potential usage of the proposed representation, we present a simple architecture for object detection. Similar to YOLO~\cite{redmon2016you}, the approach is a unified framework based on a single fully convolutional architecture. However, unlike most of other detection works, we do not use bounding box and the output is distributions representations of objects. We also do not use region proposals or anchor boxes which are proved to be useful in many works, because the proposed model aims at handling various cases in object detection and thus not making any assumptions about object shape and size. We also extend the model to output dense object masks for cases when fine object shape is needed.

\subsection{Feature Extraction and Semantic Prediction}

The whole model architecture can be viewed as a natural extension of any
semantic segmentation model, which gives dense prediction on every pixel. For
easy implementation, we adopt the DeeplabV3+~\cite{chen2018encoder} as the base
model in this work, which uses the Xception~\cite{chollet2017xception} model
backbone as the feature extractor. We keep the Deeplab segmentation branch
as-is, and extend another branch for class-agnostic object representation
prediction on the shared feature map. 

\subsection{Representation Modeling with Mixture Density Networks}

By using bivariate normal distribution $\mathcal{N}_2(\boldsymbol{\mu}_i,
\boldsymbol{\Sigma}_i)$ to represent object $i$, we need to model the five
parameters $\mu_{x_i}$, $\mu_{y_i}$, $\sigma_{x_i}$, $\sigma_{y_i}$, $\rho_{i}$.
In order to make the prediction location-invariant, for each pixel $(m,n)$ that
belongs to object $i$, the model predicts $m - \mu_{x_i}$, $n - \mu_{y_i}$,
$\log\sigma_{x_i}$, $\log\sigma_{y_i}$, $\tanh^{-1}\rho_{i}$ to form the
distribution. This can be interpreted as, at each pixel, the model is estimating
the relative location, shape, size, and rotation of the object it belongs to.
The objective function is to minimize the symmetrified KL Divergence of
predicted distribution and ground truth distribution, which is fully
differentiable. 

Although one can directly model the above characteristics for each pixel using a
single convolution layer, a potential issue is the cost of computation if the
image resolution is high. Following the common approach for dense pixel
prediction, we consider having a down-scaled prediction and up-scale it
afterward. However, common up-scaling techniques (bilinear, nearest neighbor,
etc.) can not deal with the distribution parameters at object boundary that are
not continuous. For example, bilinear upscaling will generate unexpected values
that may be viewed as parameters of another distribution by averaging parameters
of two real object distributions, resulting in predicting the boundary as
another object. In order to solve this problem, we take idea from Mixture
Density Networks to model the target distribution with $n$ distribution
candidates. For object $i$, the target distribution is modeled as
\begin{equation}
  \mathcal{N}_2(\boldsymbol{\hat{\mu}}_i, \boldsymbol{\hat{\Sigma}}_i) 
  = \mathcal{N}_2(\boldsymbol{\mu}_j, \boldsymbol{\Sigma}_j)
\end{equation}
\begin{equation}
  j = \arg\max_{k}{p_k}, k \in{1,2,...,n}
\end{equation}
where $p_k$ is the likelihood value assigned to each distribution candidate. We
use the last convolution layer to predict $6n$ parameters, which are $5$
distribution parameters and $1$ likelihood parameter for each of $n$
distribution candidates. The intuition is to let the model predict multiple
possible objects, and assign the pixel to the object with top likelihood. By
converting a value regression problem to a classification problem, the object
representation branch also fits the behavior of the base semantic segmentation
architecture.

\subsection{Optimization}

During training, the global loss function $l$ consists of three parts: the
semantic segmentation loss $l_{seg}$, the representation loss $l_{rep}$, and the
mixture density loss $l_{mix}$. 
\begin{equation}
  l = l_{seg} + \alpha l_{rep} + \beta l_{mix}
\end{equation}
where $\alpha$ and $\beta$ are two weight parameters to balance the
optimization. The parameters are related to the complexity of the scene, namely
for lower proportion of foreground objects to background, one may select higher
value for $\alpha$ and $\beta$.

For $l_{seg}$, we follow the common practice for semantic segmentation and use
per-pixel categorical cross-entropy loss on all the classes, with a binary mask
that ignores the void classes and unlabeled regions.

$l_{rep}$ is calculated using the symmetrified KL divergence $D_{\mathrm{KL}^2}$
(eq.~\ref{eq:kl}) between predicted distribution representation and the ground
truth distribution representation. Since the predicted representation is
selected from $n$ candidates in mixture density networks, we use a dynamic mask
that select two candidates at each pixel to optimize: the one with highest
likelihood and the one with lowest divergence,
\begin{equation}
  l_{rep} = \sum_i \sum_j^n D_{\mathrm{KL}^2}(\mathbb{Z}^{(i)} \parallel \mathbb{Z}^{(j)}) \cdot m_{rep}
\end{equation} 
for $\mathbb{Z}^{(i)}$ being the ground truth distribution representation at
pixel $i$, $j$ being one of the candidates prediction at pixel $i$. $m_{rep}$ is
the binary mask that $m_{rep} = 1$ if $j = \arg\min
D_{\mathrm{KL}^2}(\mathbb{Z}^{(i)} \parallel \mathbb{Z}^{(j)})$ or $j = \arg\max
p_k$ where $p_k$ is the likelihood value for candidate $k$ at pixel $i$, and
$m_{rep} = 0$ otherwise.

Lastly, $l_{mix}$ is also a categorical cross-entropy loss on the likelihood of
distribution candidates. The ground truth for which candidate being the best
option is dynamically selected as the one with lowest divergence, $j = \arg\min
D_{\mathrm{KL}^2}(\mathbb{Z}^{(i)} \parallel \mathbb{Z}^{(j)})$. Generally
speaking, we let the mixture density network automatically optimize to find the
best candidate based on its current state, and also jointly optimize the best
candidate and currently selected candidate to be close to the gound truth.
Pixels that does not belong to an object are ignored for $l_{rep}$ and
$l_{mix}$.

\subsection{Divergence-based Non-Max Suppression}

Having the dense prediction of distribution representation of potential objects,
we modify the non-max suppression to use symmetrified KL divergence
$D_{\mathrm{KL}^2}$ in place of IoU, and use it to remove false positive
detections and get the detected objects as their distribution representations. In practice, we find the threshold in invariant to object size, but variant to object class. We thus use class-dependent divergence where the class prediction is obtained from the semantic prediction.

\subsection{Instance Segmentation with Pixel Clustering}

Since all the pixels are predicted with a class label and an object
representation, we simply cluster the pixels predicted as foreground object
classes into different instances using nearest-neighbor, based on the divergence
between the pixels and detected objects after Non-maximum Suppression. For best
practice, we get object candidates on the down-scaled prediction for speed, and
cluster the pixels on original scale for accuracy. One thing to point out is
that the proposed algorithm aims to perform detection (which does not need fine
object mask), but can output object masks as an alternative for evaluation purpose. 

\section{Experiments}

In this section, we describe the experimental results quantitatively and
qualitatively explored on the selected Cityscapes~\cite{cordts2016cityscapes}
dataset. Cityscapes features 5000 images of ego-centric driving scene in urban
area, which are split into 2975, 500 and 1525 for training, validation and
testing, respectively. The ground truth contains 8 classes for foreground object
(things) with instance-level annotation, and 11 classes for background (stuff).
The reason why we choose to use this dataset is it covers a greater variety of
scene complexity and to have a higher portion of highly complex scenes than
other datasets~\cite{geiger2012we, lin2014microsoft}. Since this work aim to
solve difficult cases in object detection such as occlusion and overlapping, it
is necessary to be able to observe such cases in the dataset. Cityscapes also
provides a similar environment as real-world applications, e.g. autonomous
driving, which makes extreme demands on system performance and reliability to
handle edge cases.

\subsection{Implementation Details}

We use the 2975 finely-annotated images for training. The model is initialized
from pretrained DeeplabV3+~\cite{chen2018encoder} semantic segmentation weights
plus randomly initialized object representation branch. The images are randomly
cropped and flipped during training. Since the semantic segmentation branch is
already well-trained, we focus on samples that have higher proportion of
instance labels with a weighted sampling. We set a learning rate at $10^{-5}$
and train for 120k iterations using a desktop machine with a single Nvidia
1080Ti GPU at batch size 1. The evaluation is done with the testing set for
instance segmentation. We use single-scale images at original resolution during
inference.

\subsection{Quantitative Results}

Since most of the previous work on object detection are evaluated by average
precision on bounding box IoU, which by nature can not handle cases when ground
truth bounding boxes are highly overlapped. In order to present comparable
quantitative results, as an alternative, we report the instance segmentation
results on Cityscapes \textit{test} in Tab.~\ref{tab:instance}. Our method
achieves competitive results to state-of-the-art instance segmentation methods,
even it is not proposed to handle dense mask prediction tasks. We would like to
highlight that our method has much better performance predicting \textit{person}
and \textit{car} classes, where the overlapping are more likely to exist. For
classes like train or bus that have much larger size than other objects, our
method suffer from obtaining the contextual information of the exact location
and size of the objects, which is a common issue with single-scale models.

\begin{table}
  \caption{Instance Segmentation Results on Cityscapes \textit{test}. We 
  compare to other methods that do not use bounding box and thus can 
  potentially handle difficult cases such as box overlapping. (*: also use 
  coarse label for training)}
  \label{tab:instance}
  \centering
  \begin{tabular}{c|cc|cccccccc}
    \toprule
    Methods & AP &AP50\%  & person & rider &car &truck& bus& train& mcycle& bicycle \\
    \midrule
    InsCut~\cite{kirillov2017instancecut} & 13.0 & 27.9 & 10.0 & 8.0 & 23.7 & 14.0 & 19.5 & 15.2 & 9.3 & 4.7\\
    DWT~\cite{bai2017deep} & 19.4 & 35.3& 15.5& 14.1& 31.5& 22.5& 27.0& 22.9& 13.9& 8.0\\
    SGN*~\cite{liu2017sgn} & 25.0 &44.9& 21.8 &20.1 &39.4& 24.8& 33.2& 30.8& 17.7& 12.4 \\
    Multi~\cite{kendall2018multi} & 19.0 & 35.9 &-&-&-&-&-&-&-&- \\
    \midrule
    Ours & 19.8 & 37.5 & 21.4&16.2&41.6&15.6&23.6&17.6&10.5&11.5\\
    \bottomrule
  \end{tabular}
\end{table}

\subsection{Qualitative Results}

\begin{figure}
  \centering
  \begin{minipage}{.2\textwidth}
    \includegraphics[width=.95\linewidth]{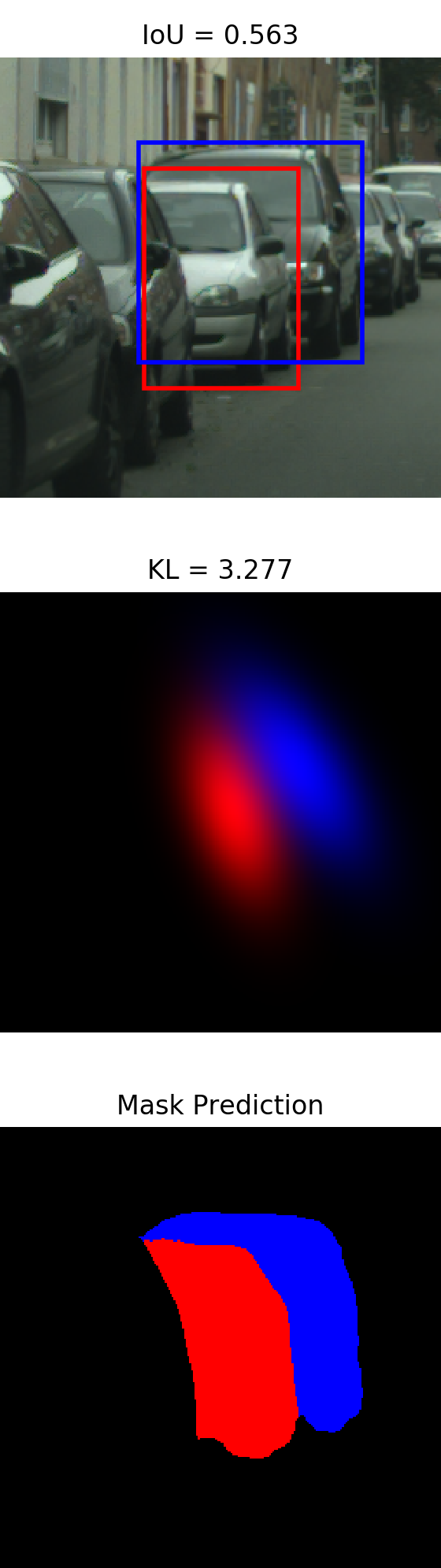}
  \end{minipage}%
  \begin{minipage}{.2\textwidth}
    \includegraphics[width=.95\linewidth]{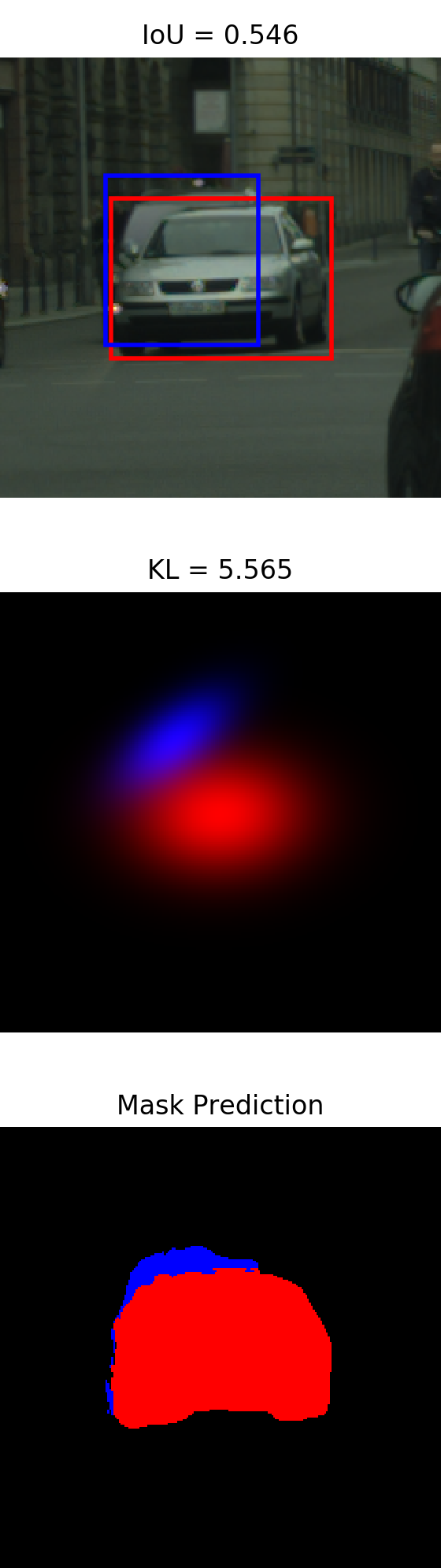}
  \end{minipage}%
  \begin{minipage}{.2\textwidth}
    \includegraphics[width=.95\linewidth]{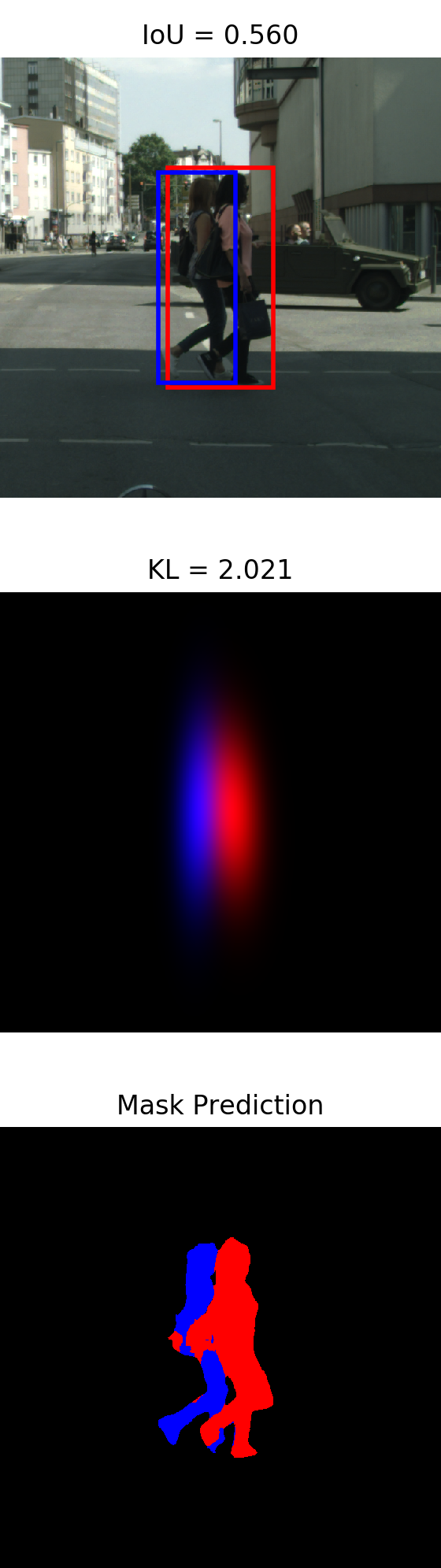}
  \end{minipage}%
  \begin{minipage}{.2\textwidth}
    \includegraphics[width=.95\linewidth]{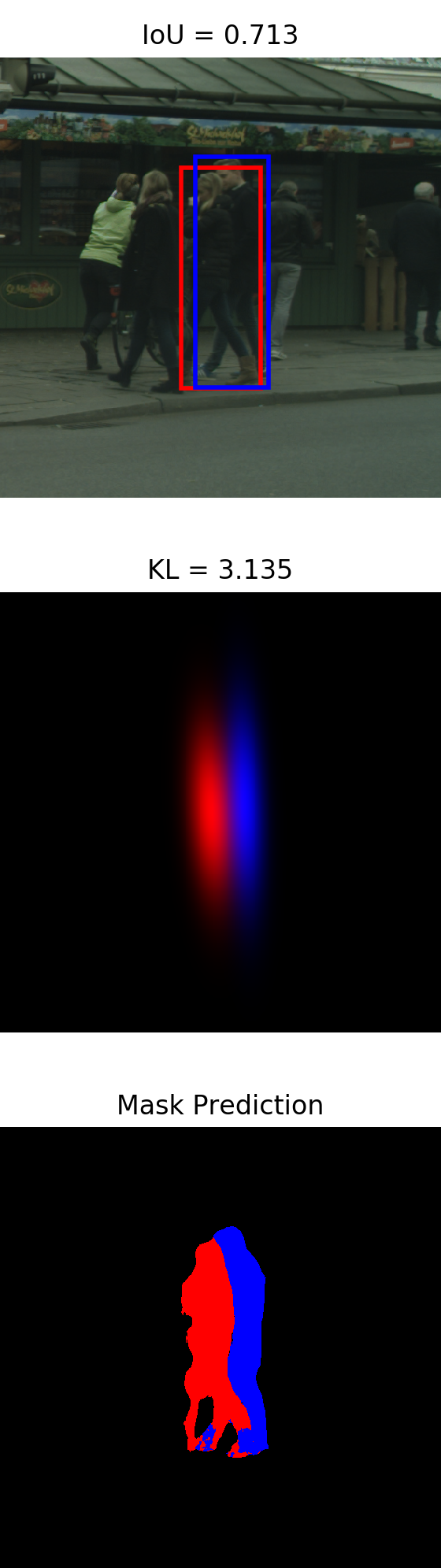}
  \end{minipage}%
  \begin{minipage}{.2\textwidth}
    \includegraphics[width=.95\linewidth]{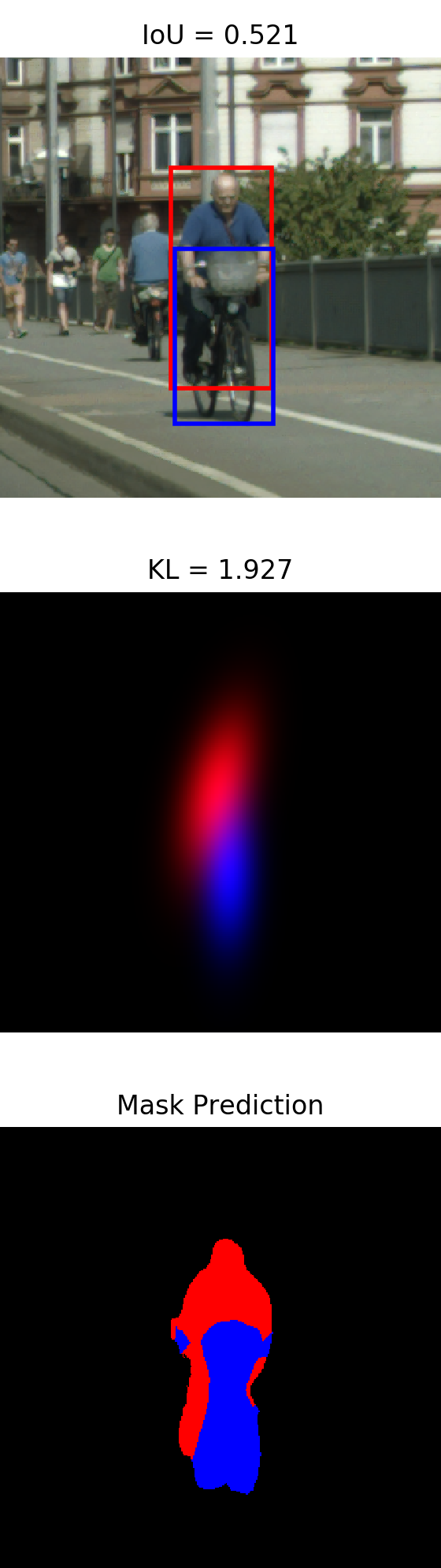}
  \end{minipage}%
  \caption{Examples of detected objects with highly-overlapped bounding boxes. From top to bottom: image crop with object bounding boxes generated from mask prediction; detection results by visualizing the predicted distribution; object mask prediction.}
  \label{fig:ioupred}
\end{figure}

We visualize the qualitative results to better illustrate the idea of proposed representation and method. Fig.~\ref{fig:ioupred} shows examples of detected objects with highly-overlapped bounding boxes, including both within-category overlapping and out-of-category overlapping situations. All these cases are very likely to be failure cases for any box-dependent methods. We claim that some of these situations can be seriously important for real-world applications, e.g. for autonomous driving, partly-occluded vehicles in the driving way are essential to path planning and collision prevention. The proposed method may also help with other related computer vision tasks such as object tracking, motion prediction, etc. 

\begin{figure}
  \begin{minipage}{.33\textwidth}
    \includegraphics[width=.98\linewidth]{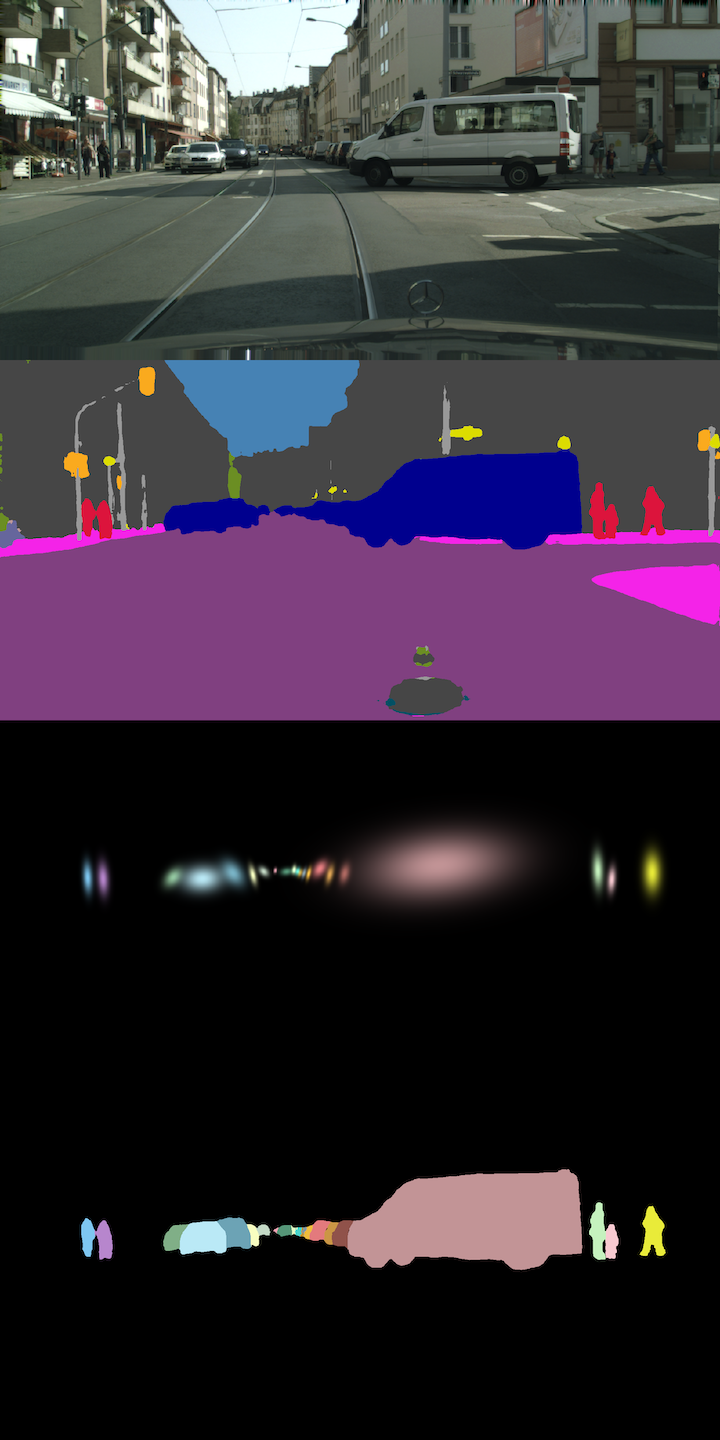}
  \end{minipage}%
  \begin{minipage}{.33\textwidth}
    \centering
    \includegraphics[width=.98\linewidth]{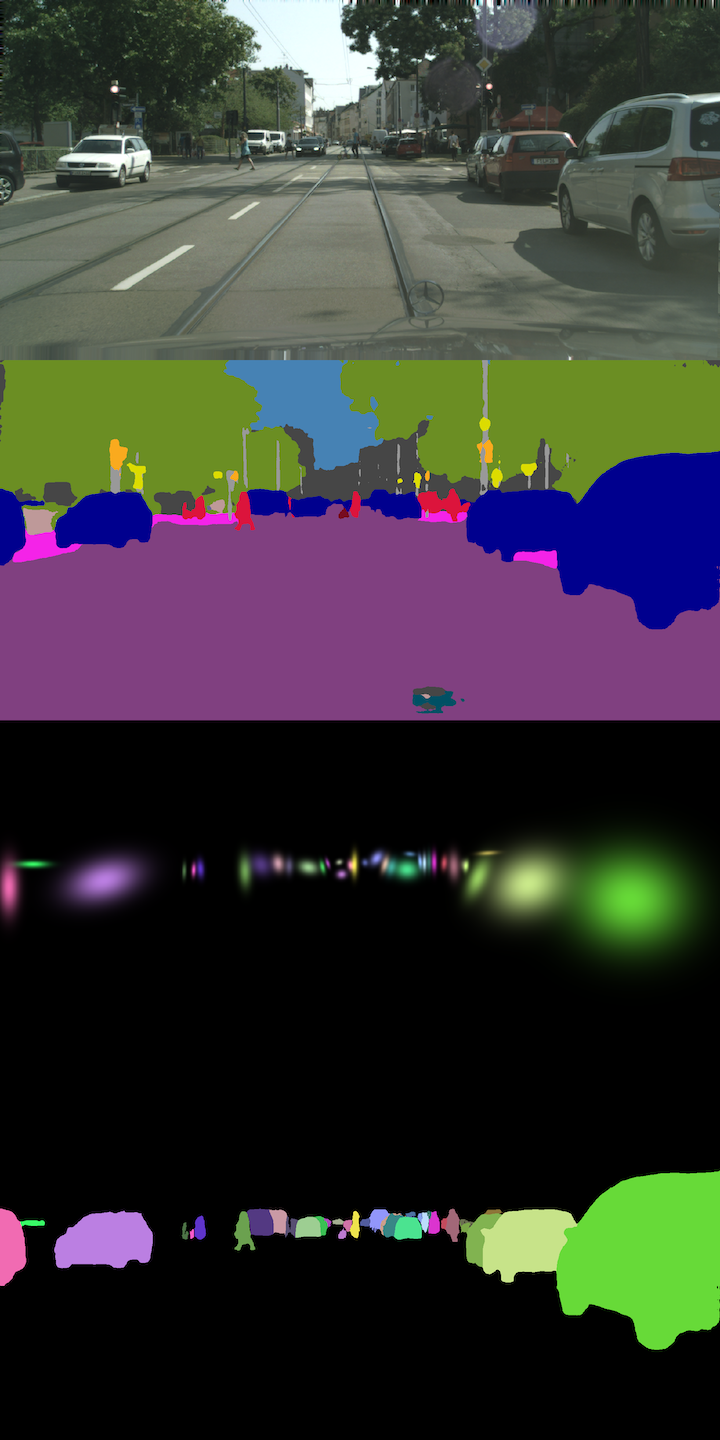}
  \end{minipage}
  \begin{minipage}{.33\textwidth}
    \includegraphics[width=.98\linewidth]{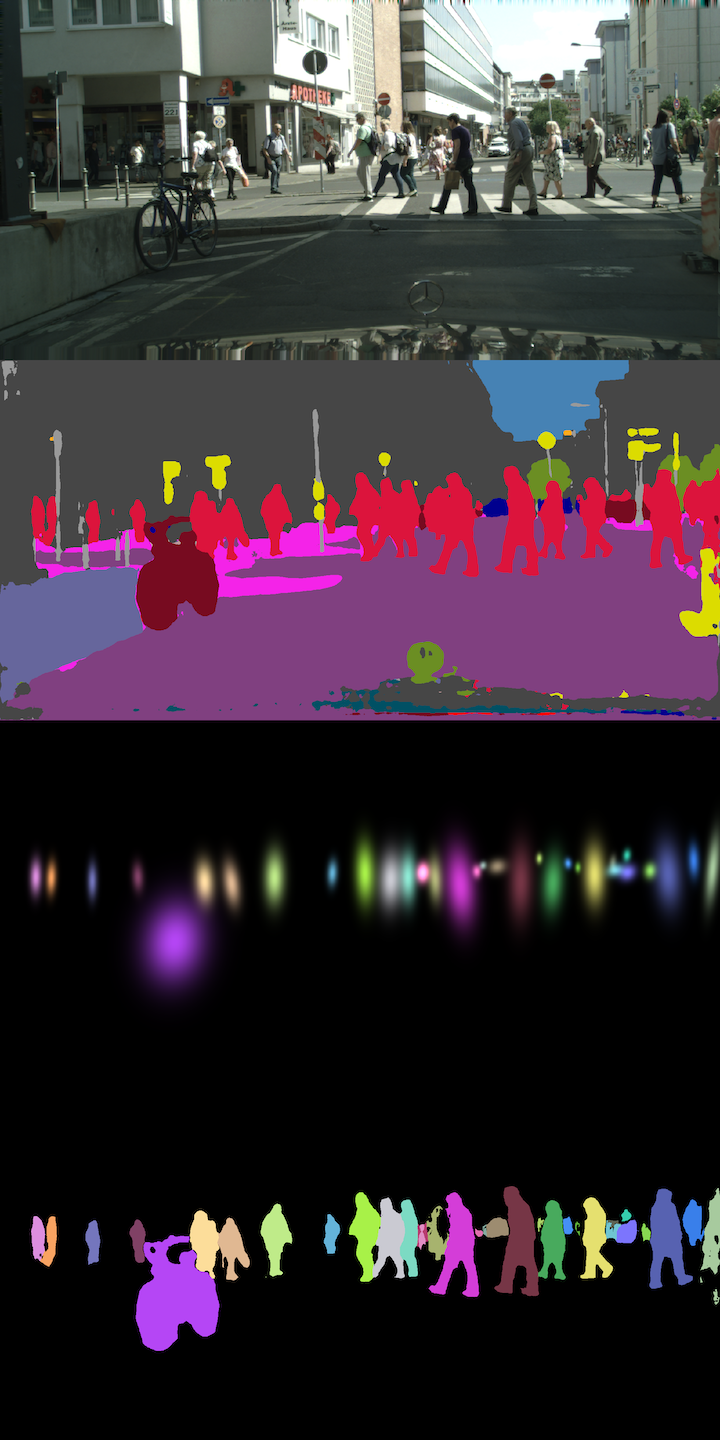}
  \end{minipage}
  \caption{Examples of prediction on various scenes. From top to bottom: input image; semantic segmentation; detection results by visualizing the predicted distribution; object mask prediction.}
  \label{fig:fullpred}
\end{figure}

We also show some visualization of model performance on different driving scenes in Fig.~\ref{fig:fullpred}. Our model provides fast and precise detection with low computational cost, and can also provide dense object masks as an extension. 

\section{Conclusion}

We propose a statistical representation of objects for the object detection task based on the bivariate normal
distribution. The qualitative evaluation shows that this representation has the benefit of robust detection of
highly-overlapping objects and the potential for improved downstream tracking and instance segmentation tasks due to the
statistical representation of object edges. 

Future work will utilize this representation for improved instance segmentation in images and temporal smoothing of both
segmentation and tracking in video. Additionally, we hope that this work raises the question whether bounding boxes is
the most useful minimalist representation of objects in real-world detection tasks such as autonomous vehicle
perception. For pedestrian, bicyclist, and vehicle detection, decoupling of overlapping objects may have elevated
significance when used as part of explicit modeling of intent and trajectory prediction.

\section*{Acknowledgment}\label{sec:acknowledgement}

Support for this research was provided by Veoneer. The views and conclusions of authors expressed herein do not
necessarily reflect those of Veoneer.

\clearpage

\small
\bibliographystyle{unsrt}
\bibliography{neurips_2019,lex}

\end{document}